\documentclass[journal]{IEEEtran}

\usepackage{xcolor,soul,framed} 

\colorlet{shadecolor}{yellow}
\usepackage[pdftex]{graphicx}
\graphicspath{{../pdf/}{../jpeg/}}
\DeclareGraphicsExtensions{.pdf,.jpeg,.png}

\usepackage[cmex10]{amsmath}
\usepackage{array}
\usepackage{mdwmath}
\usepackage{mdwtab}
\usepackage{eqparbox}
\usepackage{url}

\begin{document}
\bstctlcite{IEEEexample:BSTcontrol}
    \title{Individual and Collective Autonomous Development}
  \author{Marco~Lippi,
      Stefano~Mariani,\\
      Matteo~Martinelli,
      and~Franco~Zambonelli
  }

\markboth{Department of Engineering Sciences and Methods, University of Modena and Reggio Emilia
}
{Unimore}

\maketitle

\begin{abstract}
    The increasing complexity and unpredictability of many ICT scenarios let us envision that future systems will have to dynamically learn how to act and adapt to face evolving situations with little or no a priori knowledge, both at the level of individual components and at the collective level. In other words, such systems should become able to autonomously develop models of themselves and of their environment. Autonomous development includes: learning models of own capabilities; learning how to act purposefully towards the achievement of specific goals; and learning how to act collectively, i.e., accounting for the presence of others. In this paper, we introduce the vision of autonomous development in ICT systems, by framing its key concepts and by illustrating suitable application domains. Then, we overview the many research areas that are contributing or can potentially contribute to the realization of the vision, and identify some key research challenges.
\end{abstract}

\begin{IEEEkeywords}
Autonomous development, sense of agency, learning, self-adaptation, self-organization.
\end{IEEEkeywords}

\IEEEpeerreviewmaketitle


\section{Introduction}\label{sec:introduction}
In their early months, human infants start experiencing with their own body, moving hands, touching objects, and interacting with people around. Such activities are part of an overall process of autonomous development (a.k.a. self-development), which lets them gradually develop cognitive and behavioral capabilities~\cite{rochat1998self}. These skills include the capability to recognize situations around, the sense of self
, the sense of agency (i.e., understanding the effect of own actions in an environment)
, the capability to act purposefully towards a goal, and some primitive social capabilities (i.e., knowing how to act in the presence of others).

Moving from humans to machines, the possibility of building ICT systems capable to autonomously develop their own mental and social models and to act purposefully in an environment, is increasingly recognized as a key challenge in many areas of artificial intelligence (AI), such as robotics~\cite{lake2017building}
, intelligent IoT~\cite{Mar21}, autonomous vehicles management~\cite{Yang21}.

Indeed, for small-scale and static scenarios, and for simple goal-oriented tasks, it is possible to ``hardwire'' a model of the environment within a system, alongside some pre-designed plans of action. However, for larger and dynamic scenarios, and for complex tasks, individual components of ICT systems should be able to autonomously (i.e., without human supervision and with little or no innate knowledge): \emph{(i)} build environmental models and continuously update them as situations evolve; \emph{(ii)} develop the capability of recognizing and modeling the effect of their own actions on the context (which variables of the environment can or cannot be directly affected by which actuators, which variables and actuators relate to each other); \emph{(iii)} learn to achieve goals on this basis and depending on the current situation; \emph{(iv)} learn how to organize and coordinate actions among multiple distributed components whenever necessary.

The ambitious idea of building systems capable of autonomous development is not new, and its opportunity has been already advocated since several years~\cite{Weng599}, sometime under the notion of ``self-aware'' computing systems~\cite{selfaware}.
However, the topic is now even more timely. Many recent research results in areas such as machine learning, causal analysis, multi-agent systems, and collective behaviors, have started shedding light on the various mechanisms that have to be involved in the process of autonomous development, hinting at the fact that the vision (at least in specific application areas) is close to become reality. Furthermore, unfolding the key concepts and mechanisms underlying autonomous development can also somewhat contribute to understand the many mental mechanisms behind artificial general intelligence~\cite{Marcus21}.

Against this background, the contribution of this paper is to frame the key concepts of autonomous development in ICT systems and to identify challenges and promising research directions. Specifically, Section~\ref{sec:framework} introduces a general conceptual framework for the (continuous and adaptive) process of autonomous development, both at the individual and at the collective level; Section~\ref{sec:applications} sketches key application scenarios; Section~\ref{sec:approaches} analyzes the most promising approaches in the area of machine learning, causal analysis, multi-agent systems, and collective adaptive systems that can contribute with fundamental building blocks towards realizing the vision of autonomous development, each \emph{per se} challenging; Section~\ref{sec:challenges} identifies additional horizontal challenges to be attacked.

\section{Conceptual Framework} \label{sec:framework}





Autonomous development, beside being the process that infants carry out during the early stages of their life~\cite{rochat1998self}, also involves any ``agent'' whenever it is incarnated in a new body and immersed in a new environment.

\begin{figure}[t]
    \centering
        \includegraphics[width=0.9\columnwidth]{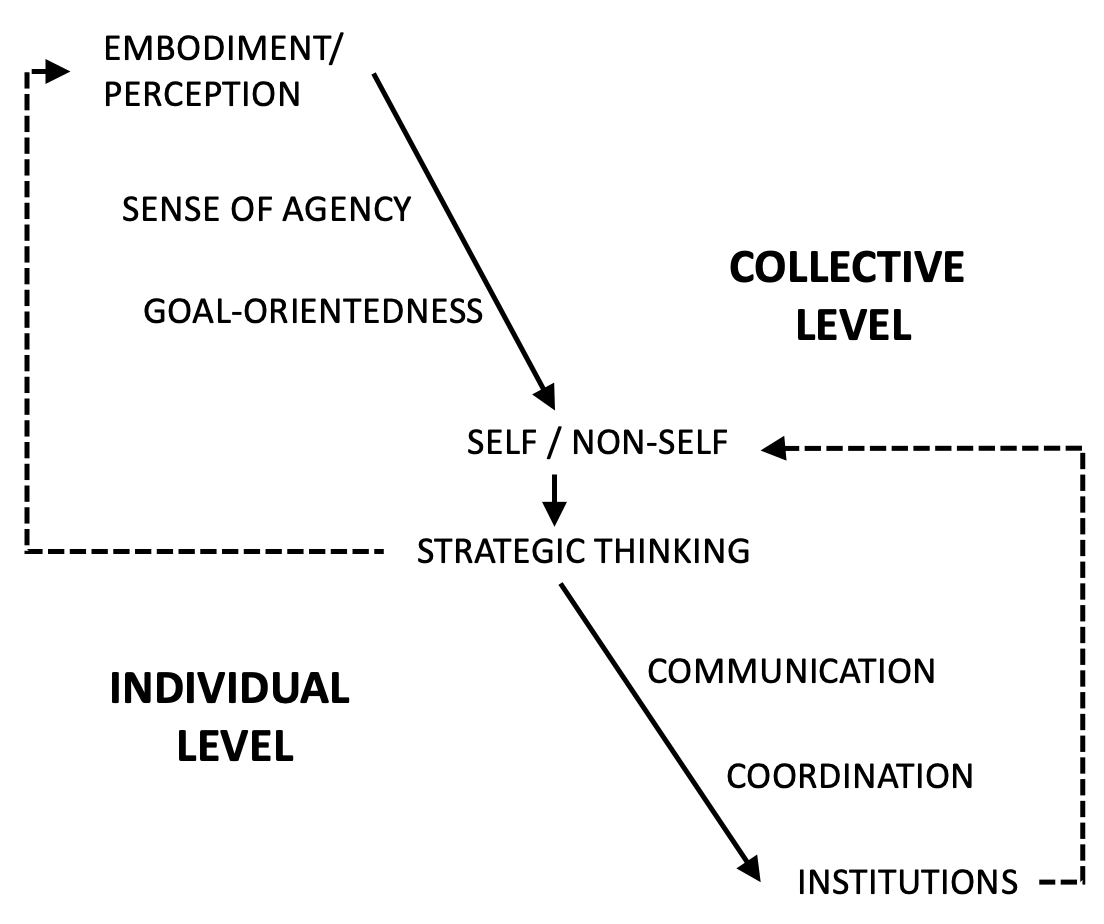}
    \caption{The conceptual framework of autonomous development.}
    \label{fig:framework}
\end{figure}

As an example to quickly and intuitively introduce our general framework (Figure~\ref{fig:framework}) let us consider what we do whenever we start playing a new video-game. At first, we observe the game environment on the screen and the commands we have available on the joystick; that is, we get acknowledged with our \emph{embodiment and perception} on the video-game. We spend a few seconds trying the commands to assess their effects in the game environment; that is, we try to acquire a \emph{sense of agency}. Then, we understand what is the goal of the game and how we can use the commands to achieve it; that is, we start acting in a \emph{goal-oriented} way.

Typically, we recognize in the video-game the presence of other ``agents'', virtual characters that are not under our control; that is, we distinguish between \emph{self and non-self}. The acquisition of such a skill implies that we acknowledge that we should tune our actions also in dependence of the actions of these other agents (\emph{strategic thinking}). All this process is typically repeated in a cyclic way (i.e., when reaching a new level in the game) to adapt to new environments, new situations, new tools available to play with, new goals, and new enemies appearing.

In the case of multiplayer games, beside recognizing the presence of players different from ourselves, and recognizing the need to act also accounting for them, we should understand: whether we have \emph{communication} tools available; how to use these tools to affect and influence the actions of others, i.e., to \emph{coordinate} with them, so that eventually \emph{institutional} ways to act together towards a goal can be established. Again, this process may be cyclically repeated as the game advances.

Truly intelligent and adaptive ICT systems should undergo a similar process and autonomously develop through similar phases. 

\subsection{The Individual Level}

Let us firstly consider a single agent $X$ (purely software or physically embodied) immersed in a (virtual or physical) environment. The agent can observe a set of environmental variables $\mathcal{V} = \{v_1, v_2, \ldots, v_m\}$. As the agent is part of the environment, internal variables of the agent itself (i.e., its current status and configuration) are included in the set. In addition, the agent has a set of actions that it can choose $\mathcal{A} = \{a_0, \ldots, a_{n-1}, null \}$, including the $null$ action. 

\vspace{3mm}

\noindent 
\emph{Embodiment and Perception}. In this very early phase, the agent should autonomously recognize the existence of $\mathcal{A}$ and $\mathcal{V}$, that is, it should get acknowledged to its actuation and sensorial capabilities. Without resorting to complex AI techniques, methods from the reflective and self-adaptive programming systems can effectively apply in this phase~\cite{sal09} to let the agent dynamically self-inspect its capabilities and start analyzing the observed variables. Still in this phase, the agent can also start acquiring some understanding of the relations between the observed variables over time, as well as some simple prediction capabilities.

\vspace{3mm}

\noindent 
\emph{Sense of Agency}. In this exploratory phase, the agent starts trying to understand what are the effects of $\mathcal{A}$ on $\mathcal{V}$, by trying to apply actions (even without any goal in mind) to see their effects. That is, it will eventually recognize that, given the current state of $\mathcal{V}_{current}$, the application of an action $a_i$ (or of a sequence of actions) will eventually lead (with some probability) to state $\mathcal{V}_{next}$. This mechanism enables the construction of the basic sense of agency~\cite{rochat1998self}, and of the sense of causality. 

\vspace{3mm}

\noindent 
\emph{Goal-orientedness}. In this exploitation phase, the agent starts applying $\mathcal{A}$ with goals in mind. That is, given the current state $\mathcal{V}_{current}$ and a desired future state $\mathcal{V}_g$ (the goal, aka the desired ``state of the affairs''), the agents resorts to the acquired sense of agency by applying the action that can possibly lead to $\mathcal{V}_g$. This also involves achieving the capability of planning the required sequence of actions to achieve a specific goal. 
    
\vspace{3mm}

\noindent 
\emph{Self and Non-Self}. As soon as an individual agent starts exploring its own actions $\mathcal{A}$, and recognizes that such actions have effect on the environment, it also understands that there are effects that are not under its own control. That is, there are ``non-self'' entities acting in the environment, too. By learning how to apply $\mathcal{A}$, the agent also learns the limits of such actions because of the non-self entities affecting some variable $v_i$.

\vspace{3mm}

\noindent 
\emph{Strategic Thinking}. The agent has built a model of the world, that is, of how $\mathcal{A}$ affects $\mathcal{V}$, and it starts including the mental models of others (non-self)~\cite{subagdja2019beyond} while acting, as well as while designing strategies.
That is, it can recognize that there are goals that it can possibly (or hopefully) attain only by accounting for the actions of others.

\vspace{3mm}

As in the video-game example, autonomous development is not to be conceived as a ``once-and-for-all'' process. Rather, it is a continuous, never-ending process: environmental conditions can change, new sensors may become available to enable more detailed observations, and new actions become feasible (or vice versa, some sensors and actions may no longer be available). This requires the agents to re-tune their learned sense of agency, and re-think how to achieve goals in isolation and in the presence of non-self entities.

\subsection{The Collective Level}

In the presence of multiple agent acting in the same environment, an agent recognizes that there are goals that cannot be achieved in isolation or by simply applying strategic thinking. Thus as part of their autonomous development, they should collectively develop some forms of ``autonomous social engagement''.

Formally, this corresponds to considering a set of $K$ agents $X_0, \ldots, X_{K-1}$, where \emph{(i)} each agent can choose the actions to perform from its own set (possibly disjoint or only partially overlapping from those of the other agents); \emph{(ii)} not necessarily all the agents can observe the whole set of environmental variables, but more likely each agent $X_j$ has the capability to perceive and/or control a subset of them. Thus, for specific goals $\mathcal{V}_g$ to be achieved, there is the need of properly combining and sequencing actions by different agents, e.g. $X_i$ executes $a^i_w$ whereas $X_j$ executes $a^j_z$, and so on.


\vspace{3mm}

\noindent 
\emph{Communication}. To overcome the limitations of strategic thinking, agents should be provided with a specific set of \emph{communication actions}, i.e., actions that are devoted to influence the actions of other agents. These could take the form of explicit communication acts, i.e., messages, that the agent should learn how to receive and send as an additional -- social -- form of perception and action. However, they could also take the form of more implicit actions aimed at affecting the behavior of others, i.e., leaving signs in the environment (stigmergy) or by adopting peculiar behaviors aimed at be noticed by others  (behavioral implicit communication)~\cite{10.1007/978-3-319-24309-2_8}. All these scenarios can be formalized by augmenting the $\mathcal{A}$ set to include communication, and possibly $\mathcal{V}$, to include observable signs in the environment. 

\vspace{3mm}

\noindent 
\emph{Coordination}. By exploring its available communication actions, agents start understanding how such acts can be used to get access and to affect some of the variables of the environment, and in particular those that are not observable and  controllable by themselves. For instance, they can learn how to use communication acts to get access to the value of some non-observable variables $v_i$ or to direct other agents in executing the actions that can affect its values as required for a goal to be achieved. In other words, such explorations enable learning basic forms of coordination, which can be thought of as a social form for the sense of agency. 

\vspace{3mm}

\noindent 
\emph{Institution}. Eventually, after exploring coordination protocols, the agents can ``institutionalise'' their patterns of interaction towards collective actions. That is, they will learn those acceptable social patterns of coordination, and the set of social norms and social incentives, that enables them to systematically achieve goals together~\cite{Morris-Martin:2019aa}. Formally, this corresponds to having agents in the collective recognize and adhere to a set of constraints $\mathcal{C(A,V)}$ ruling the way (communication) actions $\mathcal{A}$ can be performed in specific conditions $\mathcal{V}$, as well as the commitments and expectations each communication action sets on the agents participating in the protocol.

\vspace{3mm}

As for the case of the individual level, the dynamics of the environment or of the agent population may require the above collective process to assume a continuous cyclic nature. 

We emphasize that communication, coordination, and institutions are not strictly necessary to promote complex goal-oriented collective actions~\cite{leibo2019autocurricula}. Nevertheless, whenever communication mechanisms are available, learning to exploit them is a natural part of the autonomous development process, and can facilitate the effectiveness of the social development. 



\section{Application Scenarios}\label{sec:applications}

There are diverse application scenarios that can potentially take advantage of systems capable of autonomous development, at the individual and/or collective level.

\subsection{Robotics}

Robotics is the area which first identified the profitability of building systems capable to autonomously build ``by experience'' a model of their own capabilities, and consequently learn to achieve specific goals. 

In general, developers can define a model of their robots at design time, and can easily wire such model directly in embedded software without necessarily having the robot learn it in autonomy. However, autonomous learning may become necessary when the robot gets damaged while in operation, thus (partially) losing its original capabilities (e.g., a four-legged robot with one leg damaged). In that case, the robot should autonomously learn a new model to understand what it can do according to its residual operational capabilities (e.g., how to walk with three legs only), and how it can re-learn to achieve goals with them \cite{Bongard1118}.

A different situation is that of modular robots~\cite{yim2002modular}, where the robot can re-arrange its shape to serve different tasks. In this case, having the developers foresee all the possible shapes that the robot can assume to serve different tasks can be time-consuming. Also, it can prevent emergent (not previously envisioned) shapes -- functional to peculiar tasks -- to be autonomously identified by the robot itself. In this case, having the modular robot try to assume a variety of forms and understand its action capabilities for the different shapes could be very useful before deployment (other than after deployment, to recover from injuries and from the loss of some of its modules). 

At the level of collective robotics (i.e., group of robots living together in an environment and having to cooperate to achieve collective task) most current approaches relies on coordination schemes defined at design time. However, also in this case it has been argued that the autonomous evolution of communication and coordination capabilities can be of fundamental importance to acquire the capability of the collective to act in unknown and dynamically changing scenarios \cite{Cambier20}.


\subsection{Smart factories}

Similarly to a collective robotic system, a complex manufacturing system can be seen as an aggregated group of components that act together in order to achieve a production goal. Beside their basic scheme of functioning, defined at design time, if one component of the manufacturing system breaks or has some unexpected behavior, the manufacturing system should ideally adapt to the new situation, so as to overcome the problem without undermining production. 

Exploring in advance all possible contingencies and hard-wiring solutions to them in the system is almost impossible. The system should rather learn autonomously how to act upon changing conditions (whether of a temporary or ultimate nature) to maintain its overall functioning. For example, the system may explore the possibility of deviating the material flow from the broken component to another one, to learn the effect of such actions, and its overall impact on the production. In doing so, a local or global production re-scheduling might be necessary, with the initiative autonomously coming from the different components of the manufacturing system, without involving the production planning office. Clearly, such capabilities of autonomous adaptation can also apply at the level of individual components of the system, whenever these do not impact other components.

The need for integrating adaptability and flexibility in manufacturing systems is explicitly recognized as a key challenge in Industry 4.0 initiatives, and some examples of agent-based production control systems -- exhibiting some limited forms of adaptivity -- can be already found, e.g., in the automotive industry~\cite{bussman2017ICMAS-Daimler}. However, these are still far from the adaptivity level that could be reached by fully-fledged autonomous development approaches.

\subsection{Smart homes}

Buildings and homes are increasingly being enriched with sensors and actuators (i.e., IoT devices, in general), to facilitate our interactions with the environment and, by monitoring our activities and habits, to increase our safety and comfort. However, such systems are typically based on design-time decisions w.r.t. deployment of devices, their interactions, and the types of services to be provided. Learning capabilities are typically limited to monitoring user activities and adapt the parameters of services (e.g., the levels of light and heating) accordingly \cite{YeDZ19}.

From the perspective of autonomous development, we envision that once IoT devices are deployed in an environment, they should be activated to autonomously explore their own individual and collective capabilities (i.e., the individual sense of agency and the impact of inter-device interactions) so as to eventually learn how they can affect the home environment and how, and apply such capabilities once users will start populating the environment. Then, the overall smart home/building system will continue to dynamically and continuously modify its functioning to adapt to the presence of  different users with different profiles, of users whose habits tend to evolve over time, or simply to react to contingencies (e.g., modifications in the number and type of available devices, or structural modifications to the environment). 

We have conducted some simple preliminary experience to show the potential feasibility of autonomous development in a simple two-rooms smart home test-bed~\cite{Mar21}. In particular, we have shown that IoT devices in a room, when left free to explore the effects of their actions, can eventually build a sound causal model of the room, and can use such model to actuate specific environmental condition in that room (e.g., closing windows and turning off lights to achieve darkness). Also, by merging the models built for the two rooms, devices can learn to cooperatively actuate house-level environmental conditions (e.g., when a curtained window connects the two adjacent rooms, they have to agree on keeping the curtain closed so as to have light in one room and darkness in the other). Exploring larger and more complex environment and different learning techniques, and in the presence of users, will give us better clues on the general applicability of the approach and its possible limitations.

\subsection{Smart cities}

Most of the considerations above for smart homes and buildings scenario can be, in theory, transferred to the larger scenario of smart cities. That is, to all those ICT and IoT systems that can get deployed to automate and regulate the activities of modern cities (e.g., mobility, energy management, garbage collection). Indeed, the need to deploy robust systems capable --  with limited design efforts and limited human intervention -- to dynamically adapt their behavior to continuously changing urban conditions is widely recognized as a key challenge for harmonic and sustainable urban development~\cite{Ullah20}.

The substantial difference between smart homes/buildings and smart cities is that cities already exist and are already inhabited. Thus, you cannot think at let a smart city system free to explore the effect of its actions and interactions to eventually become capable to act in a goal-oriented way (which you can instead do before a new building becomes inhabited).

However, one can think at exploiting a simulation-based approach to this purpose. Given that accurate and reliable simulators exists to study different urban aspects, one can think at having system components explore and learn in a simulated environment towards full mental development, before being eventually deployed in the real world \cite{Yang21}. Indeed, exploiting simulators to support autonomous development, the same as children are given toys and playgrounds to explore and develop their capabilities, can be a key for any application area.


\section{Research Approaches}\label{sec:approaches}


\begin{figure*}[t]
    \centering
        \includegraphics[width=0.75\textwidth]{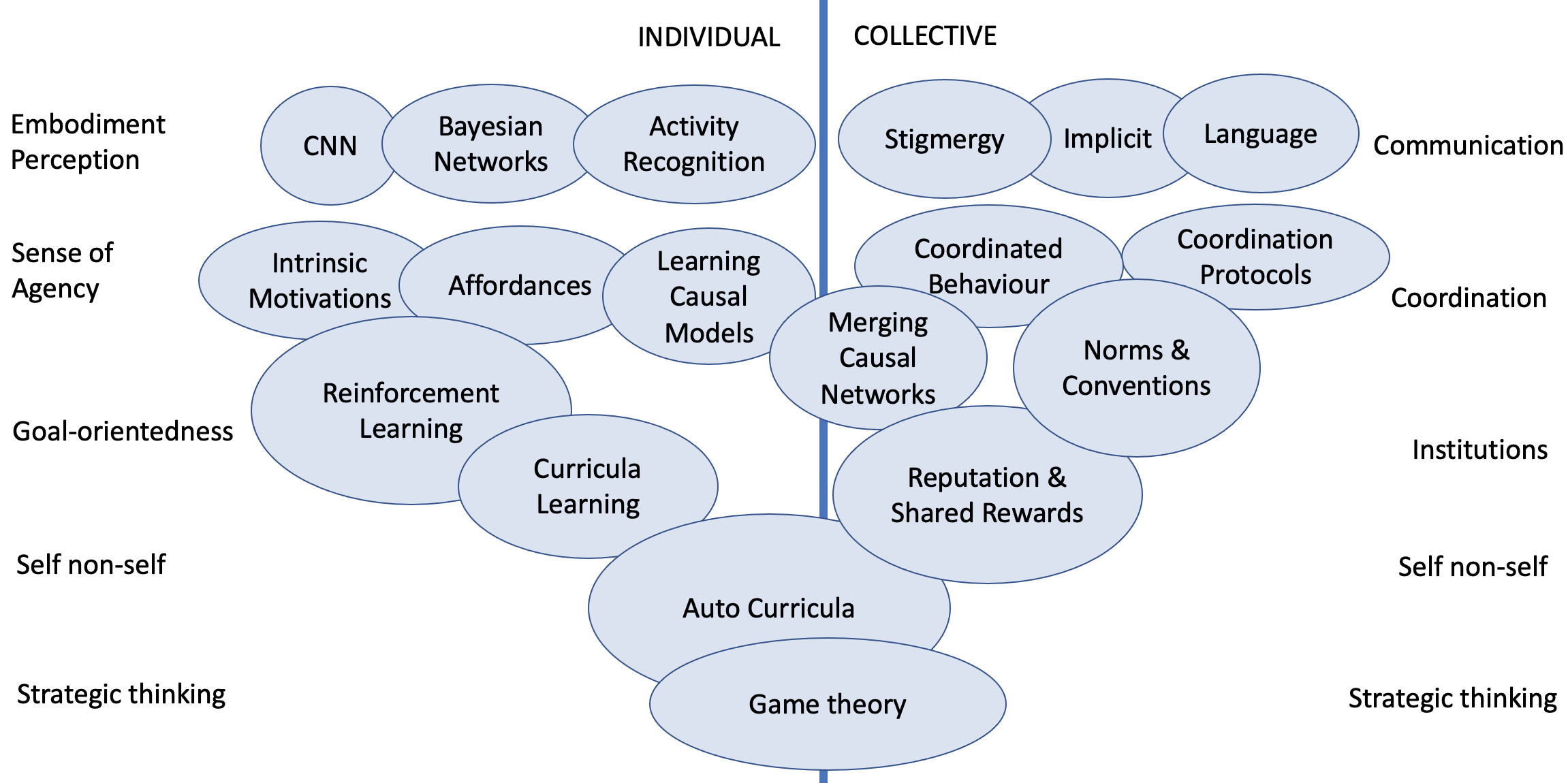}
    \caption{The galaxy of autonomous development.}
    \label{fig:approaches}
\end{figure*}

The idea of autonomous development, at both the individual and collective level, has been widely investigated in areas such as cognitive psychology, neuroscience, philosophy, and ethics~\cite{Weng599}. 
We hereby focus on the computational perspective, and in particular on the galaxy of different approaches that can contribute to unfold the mechanisms involved in autonomous development vision and to eventually realize the vision (Figure~\ref{fig:approaches}). Although most of these approaches can play or are already playing a fundamental role, they still have to attack several challenging problems to become practical tools for future systems engineering. 

We do not focus here on the basic levels of individual autonomous development, i.e., perception and embodiment, in that tools already exist to give agents sophisticated sensing abilities (e.g., convolutional neural networks to recognize objects, scenes, and activities) and the capability of controlling their own actuators purposefully. 

\subsection{Goal-oriented Learning}

The broad area of reinforcement learning shares with our vision the objective of training machines to act in a goal-oriented way in a specific context. However, despite the amazing recent results in the area, in particular with deep Q-learning~\cite{mnih2015human}, most current approaches do not aim at building systems with a sense of agency and capable of developing an interpretable world model, but rather at achieving goals based on explicit, domain-based rewards, that are named \textit{extrinsic}. This makes most approaches highly ineffective in scaling up to learning tasks in complex contexts, or across domains, or despite the ever-changing dynamics of the environment.

Curriculum-based approaches to machine learning go somewhat in the direction of gradually developing the capability to act in complex scenarios~\cite{bengio2009curriculum}. The agent is first trained on simple tasks, and the gained knowledge is accumulated and exploited in increasingly complex scenarios, where further skills can thus be effectively learnt. Yet again, most of these approaches do not focus on the development of a world model and of an explicit sense of agency.

Reinforcement learning approaches based on \emph{intrinsic} rewards~\cite{Sch10}
, instead, more closely exploit the idea of exploring the world to develop a sense of agency. In fact, while extrinsic rewards are typically designed by a ``teacher'' (e.g., the score in a videogame) intrinsic rewards are developed by the agent itself to satisfy its curiosity (i.e., when it discovers how to achieve specific tasks). For example, in~\cite{burda2018large} intrinsic rewards are computed as the error in forecasting the consequence of the action performed by the agent given its current state. 

Recent approaches based on the theory of affordances~\cite{Khet20} propose to have agents gradually learn the effects of their actions. By having them act in constrained environments where only a limited set of actions apply, they eventually develop an explicit sense of agency, i.e., a model of how their actions affect the environment. 

Al these approaches face the key challenge of building general conceptual and practical tools to: (i) learn to effectively act in an environment by exploiting the power of model-free sub-symbolic (deep learning) approaches; and, at the same time, (ii) learn incremental and reusable, possibly causal models of the world. The latter being increasingly recognized as a key ingredient for intelligence and autonomous development, whereas they mostly still lack effort and success is in building \emph{explicit, actionable models} of the world, there including the environment and the society of agents.

\subsection{Learning causality}

Understanding and leveraging \textit{causality} is recognized as a key general challenge for AI in the coming years~\cite{scholkopf2021toward}. In particular, Judea Pearl~\cite{pearl2019} 
has proposed the idea of a ``causal hierarchy'' (also named ``ladder of causation'') to define different levels of causality recognition and exploitation by an intelligent agent. 

The first level of the ladder consists in simply detecting relations as associations, whereas the second one assumes the possibility to intervene in the environment and observe the (causal) effects of the taken actions. Finally, the third level enables reasoning and planning on the basis of counterfactual analysis. Such layers correspond to some of the phases of the autonomous development loop we defined: the first one is mostly involved in the perception phase, whereas the second one is associated to the development of a sense of agency and to recognition of self and non-self. The final layer clearly enables goal-oriented behavior, strategic thinking, and collective coordination.

Bayesian and causal networks are among the models that are most widely exploited in order to build interpretable causal models of the world. A recent contribution that is in line with the ideas we envision for autonomous development is the application of curriculum learning to the problem of learning the structure of Bayesian networks~\cite{zhao2017learning}. On a pure sub-symbolic level, on the other hand, another recent work proposes to learn causal models in an online setting~\cite{javed2020learning}, with the aim to find (and strengthen) causal links between input and output variables. 

We argue that key challenges in this area concern, again, understanding how to synergistically exploit symbolic and sub-symbolic approaches to learn, represent, and evolve causal models in autonomous development scenarios, and how to use them to adaptively achieve goals.




\subsection{Autocurricula}

When multiple agents act in a shared environment, their actions and their effectiveness in achieving goals are affected by what others do. Game-theoretic approaches to strategic thinking have deeply investigated this problem and the decision-making processes behind \cite{myerson2013game}. In this context, it has also been shown that agents can effectively learn in autonomy to improve their performance in dealing with others~\cite{nowe2012game}.

However, when moving from theoretical settings (e.g., the prisoner's dilemma) to complex and realistic scenarios where agents have complex goals, peculiar phenomena arise. The more one agent learns, the more it challenges others, triggering a continuous increase in complexity of behaviour, ultimately enabling to incrementally learn more sophisticated means to act. This somewhat resembles the increase of complexity that agents face in curricula approaches to reinforcement learning. The key difference being that, in the presence of multiple agents, the increase in complexity and capabilities of agents is promoted and self-sustained by the system itself, hence the term \emph{autocurricula}~\cite{leibo2019autocurricula}. 

Recently, autocurricula-based approaches have produced stunning results in multiagent environments, both cooperative and competitive. For instance, in a hide and seek scenario \cite{baker2020emergent}), agents moving in a complex simulated environment have learned how to effectively compete (hiders against seekers) and cooperate (coalitions of cooperating seekers/hiders) in very elaborated ways, in a continuous self-sustained learning process. Indeed, we consider such approaches fundamental towards the autonomous development of complex agent societies. Yet, a deep understanding of the process that drives evolution of individual and collective behaviors is still missing, and is a key challenge for the next few years. To this end, providing agents an \emph{explicit} modeling (possibly in \emph{causal} terms) of the others' behavior and of the overall societal behavior, may be necessary~\cite{subagdja2019beyond}. Also, autocurricula approaches do not currently account for the possibility of explicitly interacting (e.g., through speech acts) with other agents, which may indeed fundamental to improve collective learning.

\subsection{Learning to communicate and coordinate}

Agents may communicate and coordinate by explicit messages, 
by leaving traces in the environment,
or implicitly~\cite{10.1007/978-3-319-24309-2_8}.

These forms of communication are already exploited in multiagent learning, mostly to improve the individual learning process by letting agents share information (e.g., for merging their individual causal models of the world ~\cite{meganck2005distributed})
) and coordinate actions.
However, these communication approaches are usually assumed as an \emph{innate} capability of agents, rather than one to be learnt. That is, agents have an \emph{a-priori} sense of agency with respect to communication actions, whereas in our vision it should be developed by learning.

For example, with reference to explicit communication acts, \cite{DBLP:conf/atal/GuptaD20} proposes a voting game to let agents learn to share a communication language and to develop a strategy to communicate. In~\cite{Foer16}, it is shown that reinforcement learning can be effectively applied to let agents learn how to communicate in order to achieve a specific effect.
%
In the case of \emph{implicit} communication, instead, forms of implicit behavioral communications have been shown to emerge in simple system components that purposefully move in an environment~\cite{Grupen20}, as they learn to affect others with \emph{ad-hoc} actions. 
Learning to use stigmergy to effectively coordinate is under-explored in the literature, which instead focuses on the opposite -- using stigmergy to boost learning.
%

In any case, the development of general approaches to let agents developed fully-fledged forms of communication and coordination is still an open challenge, which may call for agents to develop not only a model of the world, but an overall model of the society, i.e., a social sense of agency explicitly modeling how communications and coordination actions affect other agents in the shared environment.


\subsection{Emergence of Institutions}

Whereas learning to communicate is about understanding how to use communication to coordinate actions with others, enabling and sustaining global collective achievement of goals requires ``institutionalized'' means of acting at the collective level, i.e., a set of shared beliefs and of shared social conventions and norms aimed at ruling collective actions \cite{Esteva01}.
The mechanisms leading to the spontaneous emergence of institutions in human society,
there included the mechanisms to promote and sustain altruistic and cooperative behavior (e.g., reputation and shared rewards),
have been widely investigated~\cite{Now06}. However, most approaches to building multiagent systems assume such mechanisms as \emph{explicitly designed}~\cite{Esteva01}. 

Yet, some promising studies related to the emergence of institutionalized behaviors in multiagent systems have been undertaken (see~\cite{Morris-Martin:2019aa} for a recent survey).
For instance, \cite{DBLP:journals/tcyb/YuZR14} proposes a collective learning framework where agents learn to adopt norms in repeated coordination, i.e., agents eventually \emph{learn} that a social norm has emerged, and ``institutionalize'' their behaviour in their (social) decision making processes \emph{implicitly}, by behaving so as to comply to the norm.
%
Another interesting work~\cite{10.5555/2615731.2616158} integrates rational thought, reinforcement learning, and social interactions to model norms emergence in a society: agents incrementally develop a social behaviour (a social norm) while \emph{internalising} it within their cognitive model.

However, the development of general models and tools to support the proper learning and evolution of institutionalized mechanisms of coordination, through the construction of explicit norms representations and adoption by agents cognitive models is still missing, and so are the solutions to the many problems involved in this process. For instance: how to avoid that an agent learns that free-riding is better than abiding to norms; or how to avoid inconsistencies and misunderstandings in their interpretation.
%

\section{Horizontal Challenges}\label{sec:challenges}



The presented approaches and techniques are still at the research stage, and many research challenges have been identified for each of them. In addition, it is possible to identify several additional ``horizontal'' challenges, i.e., of a general nature independently of the specific approach.

The specific nature of such challenges, in our opinion, makes them specifically suited for being pursued by the most diverse research communities in the area of machine learning, reinforcement learning, autonomous and multi-agent systems. 
%
\vspace{3mm}

\noindent 
\emph{Engineering}. Many of the presented approaches are grounded in the area of machine learning and multi-agent systems, disciplines with plenty of years of research behind, but in which traditional software engineering problems are sometimes considered mundane. Systems are often developed \emph{ad-hoc} for a specific task or problem domain, with little attention to modularity, re-usability, dependability, thus missing the flexibility to adopt them across different domains, tasks, data-sets~\cite{porter2019distributed}. In addition, given that the diverse approaches presented can each contribute important pieces to the overall vision of autonomous development, sound engineering approaches are needed to try to integrate such a heterogeneous plethora into a coherent whole. These represent multi-faceted and horizontal research challenges that, in our opinion, could and should be profitably attacked by promoting synergies and collaboration with the software engineering research community.


\vspace{3mm}

\noindent 
\emph{Controlling evolution}. Autonomous development raises the issue of somewhat controlling how  behaviors evolve, as individual learns new skills and tasks, and as the collective learns new way of coordinating and acting together. How can we \emph{steer} a learning process towards desired outcomes without putting bias in it? How can we \emph{constrain} the boundaries within which individual and collective behaviors should stay (e.g., in terms of safety)? What \emph{interventions} can we make to re-direct an agent or a collective that has taken an unpredictable or unsafe autonomous development path? Experience in self-adaptive components based on feedback, as well as in the study of emergent behaviors in self-organizing systems and definitely help in finding proper technical answers, and -- why not -- \emph{ethical} ones~\cite{DBLP:conf/hicss/YapoW18}.

\vspace{3mm}

\noindent 
\emph{Humans in the Loop}. The more technologies based on autonomous development will advance, the more humans will have to actively interact with them. This interaction will raise technical issues (will we have ``handles'' to control or block such systems in some ways and to some extent?) and ethical problems (will we be rather ``handled'' by these systems and subjects to their decisions?). Some of these problems already emerged, like in the \textit{moral machine} experiment~\cite{awad2018moral} or in AI-based hiring technology.
%
%
Technical challenges will be meat for the HCI and distributed systems communities (there included the self-organizing systems one). Ethical and moral ones will be meat for politicians and lawyers, although deep joint work with technical experts will always be necessary.
%
%
%
%
%
%
A key ingredient involves institutions, since they represent humans as a group: laws and regulations need to be developed to regulate the global actors into the day-by-day technology usage. Nevertheless, a deeper interaction between researchers in science and technology and public institutions is needed to support the regulation design phase.
%


\vspace{3mm}

\noindent 
\emph{Sustainability}. Algorithms for autonomous development will most likely require extensive computational resources. For example, the mentioned ``hide and seek' experiment by OpenAI involved a distributed infrastructure of 128,000 pre-emptible CPU cores and 256 GPUs on GCP~\cite{baker2020emergent}: the default model optimized over 1.6 million parameters taking 34 hours to reach the fourth stage over six of agents skills progression. 
%
This example is a sort of best-in-class projects; anyway, it is clear that if autonomously developing systems will be based on similar learning approaches, they will require massive amounts of computational resources. 
%
%
Therefore, a key challenge for the community will be to devise algorithmic and system-level means to make autonomous development systems sustainable, and affordable by others other then the big technology players.
%



\vspace{3mm}

\noindent \emph{Explainability}. Being able to inspect and explain the decision making process of AI systems is already a hot topic, so much that an entire research field (XAI, from eXplainable AI) has born. We already commented several times how such problems should be compulsory accounted for also for autonomous development, possibly with the help of causal models or other symbolic models. This is indeed a key challenge for the several research research community, too, where explaining individual and social behaviors, and patterns and global configurations emerging from local interactions is mostly still considered a ``holy grail''.

\section{Conclusions}\label{sec:conclusions}

The general vision of autonomous development is still far to be reality. However, several ideas in the areas of machine learning, causality, multi-agent systems, are already showing its potential feasibility and applicability, at least in specific application areas. We argue that researchers in these areas have plenty of room for further exploring the topic and contribute to advance the vision, possibly addressing the many open issues that we have tried to identify.



%





\ifCLASSOPTIONcaptionsoff
  \newpage
\fi





\bibliographystyle{IEEEtran}
\bibliography{biblio}

\vfill


\end{document}